\documentclass[journal]{IEEEtran}
\usepackage{amsmath,amssymb,amsfonts}
\usepackage{algorithmic}
\usepackage{graphicx}
\usepackage{textcomp}
\usepackage{xcolor}
\PassOptionsToPackage{prologue,dvipsnames}{xcolor}

\usepackage{array}
\usepackage{multirow}
\usepackage{rotating}
\usepackage{ragged2e}
\usepackage{booktabs}

\usepackage{stfloats}
\usepackage{url}
\usepackage{verbatim}
\usepackage{cite}
\usepackage[normalem]{ulem}
\usepackage{tikz}
\usepackage{threeparttable}  
\usepackage[caption=false,font=normalsize,labelfont=sf,textfont=sf]{subfig}
\newcommand{\model}{HiGNN}

\let\b\boldsymbol
\usepackage{commath}
\usepackage{bbm}

\newtheorem{definition}{Definition}
\newtheorem{theorem}{Theorem}[section]

\ifCLASSINFOpdf
\else
\fi
\usepackage{amsmath}
\usepackage{array}

\usepackage{stfloats}
\usepackage{url}

\begin{document}
%
\title{Learn from Heterophily: Heterophilous Information-enhanced Graph Neural Network}
%
%
%

\author{Yilun Zheng, Jiahao Xu, 
       Lihui Chen,~\IEEEmembership{Senior Member,~IEEE}
\thanks{Yilun Zheng, Jiahao Xu, and
       Lihui Chen are with the Centre for Info. Sciences and Systems, Nanyang Technological University, Singapore 639798, e-mail: yilun001@e.ntu.edu.sg; jiahao004@e.ntu.edu.sg; elhchen@ntu.edu.sg}
}

%
%

\markboth{Journal of \LaTeX\ Class Files,~Vol.~14, No.~8, August~2021}%
{Shell \MakeLowercase{\textit{et al.}}: A Sample Article Using IEEEtran.cls for IEEE Journals}


%




\maketitle

\begin{abstract}
Under circumstances of heterophily, where nodes with different labels tend to be connected based on semantic meanings, Graph Neural Networks (GNNs) often exhibit suboptimal performance. Current studies on graph heterophily mainly focus on aggregation calibration or neighbor extension and address the heterophily issue by utilizing node features or structural information to improve GNN representations. In this paper, we propose and demonstrate that the valuable semantic information inherent in heterophily can be utilized effectively in graph learning by investigating the distribution of neighbors for each individual node within the graph. The theoretical analysis is carried out to demonstrate the efficacy of the idea in enhancing graph learning. Based on this analysis, we propose HiGNN, an innovative approach that constructs an additional new graph structure, that integrates heterophilous information by leveraging neighbor distribution to enhance connectivity between nodes that share similar semantic characteristics. We conduct empirical assessments on node classification tasks using both homophilous and heterophilous benchmark datasets and compare HiGNN to popular GNN baselines and SoTA methods, confirming the effectiveness in improving graph representations. In addition, by incorporating heterophilous information, we demonstrate a notable enhancement in classicial GNN baselines across real-world datasets, thus affirming the efficacy of our approach. 
\end{abstract}


%
\IEEEpeerreviewmaketitle

%
%
%
%

\section{Introduction}

\IEEEPARstart{G}{raph-structured} data widely exists in various applications including social networks\cite{gnn_social_nn} and recommendation systems\cite{gnn_recom_nn} where entities are connected by their relations. To learn representations from these relational graphs, Graph Neural Networks(GNNs)\cite{GCN, GAT,graphSage, yang2022ncgnn} are proposed and have made substantial advancements. The effectiveness of GNNs on graph-structured data largely hinges on the homophily assumption, which posits that connected nodes tend to be similar\cite{rw_hom_asump,zhao2022neighborhood,wu2023beyond,wu2023homophily,chen2023exploiting}. Under this assumption, information is aggregated from neighboring nodes with similar features, facilitating effective information exchange within local structures. However, the homophily assumption does not always hold in many real-world scenarios where dissimilar nodes are connected\cite{hom_not_hold}, a situation referred to as heterophily or non-homophily. Heterophily poses a limitation to the performance of GNNs as it involves the aggregation of information from dissimilar nodes, thereby introducing additional noise during the aggregation process.

To handle the heterophily in graphs, several approaches have been proposed. Some methods calibrate the aggregation process by integrating high-order neighbors with different weights\cite{Mixhop,H2GCN,GPR-GNN,HOG-GCN}, introducing adjacency matrices as a new feature of nodes\cite{LINKX}, or propagating prior belief estimations\cite{CPGNN}. While these aggregation calibration methods have demonstrated some improvements on heterophily graphs, they primarily focus on local neighbors and fail to aggregate information from global neighbors. Consequently, alternative methods are proposed to aggregate signals from global neighbors by integrating new filters\cite{FAGCN,ACM-GCN}, or construct new graph structures\cite{Geom-GNN,GGCN,WRGAT,GloGNN}. 

\begin{figure}
    \centering
    \includegraphics[width=0.5\textwidth]{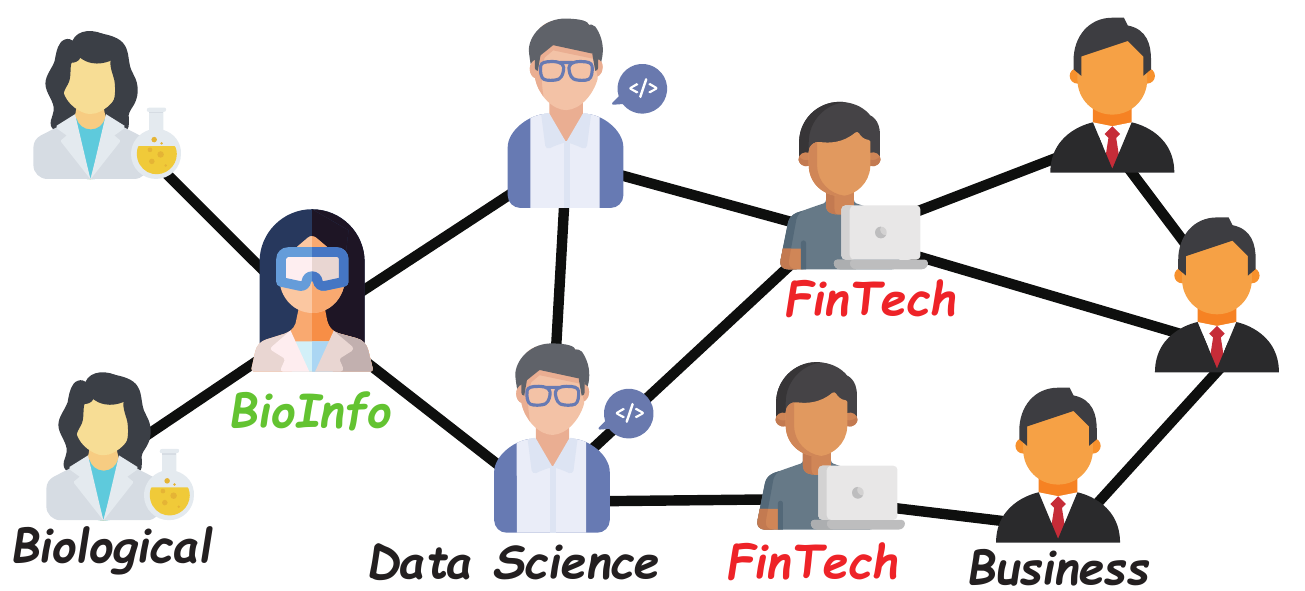}
    \caption{An academic collaboration graph with many interdisciplinary collaborations, such as scholars from Bioinformatics(BioInfo) and Financial Technology(FinTech), resulting in high heterophily with rich semantic information.}
    \label{fig:Hi_example}
\end{figure}

Although the heterophily is conventionally considered as noises in many graph-based machine learning tasks during message aggregation\cite{ACM-GCN,graphhom_gcls,graphhom_lp}, there are scenarios where a heterophilous graph structure could carry meaningful semantic information. For example, as an academic collaboration network shown in Figure \ref{fig:Hi_example}, scholars from different disciplines are connected by academic collaborations. The heterophily in this network reflects meaningful interdisciplinary collaborations, such as Bioinformatics (collaborations between Biology and Data Science) and Financial Technology (collaborations between Business and Data Science). In this case, a high degree of heterophily indicates a rich exchange of ideas and methodologies between different fields, fostering innovation and broadening research perspectives. While these heterophilous graphs contain rich semantic information, the performance of GNNs is limited by the heterophily. This leads to a crucial question: How can we utilize useful semantic information from the heterophily?

To extract the semantic information from the heterophily, we first define the conception of heterophilous information, which describes the probability of a node’s neighbors belonging to specific classes. This information not only provides insight into the degree of homophily or heterophily but also enhances our understanding of the heterophilous contexts of nodes. As the previous example shown in Figure \ref{fig:Hi_example}, scholars in Bioinformatics or Financial Technology share a high degree of heterophily with their neighbors. Although these heterophilous contexts are conventionally regarded as noise during message aggregation in GNNs, their heterophilous information can identify them as interdisciplinary collaborators in specific research areas. To better understand how effective the heterophilous information is in improving GNN performance, we conduct theoretical and empirical analyses to assess its impact. The results show that heterophilous information is effective in identifying semantically similar nodes. Building upon this analysis, we propose a novel method, \textbf{\model{}} (\textbf{H}eterophilous \textbf{I}nformation-enhanced \textbf{G}raph \textbf{N}eural \textbf{N}etwork), designed to incorporate heterophilous information into GNNs. Leveraging this heterophilous information, the \model{} demonstrates superior performance on both homophilous and heterophilous datasets, significantly improving the homophily degree within newly constructed graph structures. In summary, our main contributions are summarized as follows:

\begin{itemize}
    \item The concept of heterophilous information in graphs is defined for the first time, and its effectiveness in improving homophily is verified through theoretical and empirical analysis.
    \item The HiGNN method is proposed, which constructs a new adjacency matrix and establishes new connections between nodes based on the similarity of heterophilous information, thus distinguishing semantic information from noises and improving the performance of graph representation at different levels of heterophily.
    \item The experimental results demonstrate the superiority of our method on both the homophilous and heterophilous datasets. We also show improvements over existing GNN baselines by introducing heterophilous information.
\end{itemize}
\section{Background}
We denote $\mathcal{G = (V,E)}$ as an undirected graph $\mathcal{G}$ with node set $\mathcal{V}$ and edge set $\mathcal{E}$. Let $A$ be an adjacency matrix where $A_{ij}=1$ or $A_{ij}=0$ represents the presence or absence of an edge $e_{ij}$ connecting node $v_i$ and $v_j$. For each node $v_i$, we use $\mathcal{N}_i = \{v_j|A_{ij}=1\}$ to denote its 1-hop neighbor set and $d_{i}$ to denote its node degree. We represent the node features as $X=\{x_1,x_2,...x_n\}\in \mathbb{R}^{N\times M}$ and the labels as $Y=\{y_1,y_2,...,y_n\}\in \mathbb{R}^{N\times 1}$, where $N$ and $M$ refer to the number of nodes and features in $\mathcal{G}$ respectively.

\textbf{Graph Neural Networks} (GNNs) are a class of neural networks designed to process graph-structured data. Based on the message passing mechanism, GNNs enable nodes to exchange information with their neighbors. In the $l$-th layer of GNN, the representation $z_i$ for node $v_i$ is updated as follows:
\begin{equation}
    z_i^{l} = \text{UPDATE}^l \left( z_i^{l-1},\text{AGGREGATE}^l(z_i^{l-1},\{z_j^{l-1}|v_j\in \mathcal{N}_i\})\right)
\end{equation}
where $\text{AGGREGATE}(\cdot)$ describes the aggregation of information from node $v_i$ and its neighbors $\mathcal{N}_i$, and $\text{UPDATE}(\cdot)$ describes how the representaion $z_i^l$ is updated based on ego representation $z_i^{l-1}$ and aggregated neighbor representations. The Graph Convolutional Network(GCN) is one of the representative of GNNs and the output of its $l$-th layer can be represented as:
\begin{equation}
    Z^{l} = \sigma(\hat{A}Z^{l-1}W^{l-1})
\end{equation}
Here an activation function $\sigma$ is applied to the normalized adjacency matrix $\hat{A}=\Tilde{D}^{-\frac{1}{2}}\Tilde{A}\Tilde{D}^{-\frac{1}{2}}$, where $\Tilde{D} = D+I_n$ and $\Tilde{A} = A+I_n$ represent the degree matrix and adjacency matrix with added self-loops. 

\textbf{Graph homophily} is the fundamental assumption of GNNs, where similar nodes are likely to be connected. This principle of homophily enhances the efficacy of information propagation within GNNs, as it aggregates the representations of nodes sharing identical labels. The commonly used metrics for graph homophily include edge homophily\cite{Mixhop} and node homophily\cite{Geom-GNN}, which can be defined as:
\begin{equation}\label{equ:def_hom}
\begin{split}
    h_{edge} &= \frac{|\{{e_{ij}|e_{ij}\in \mathcal{E}, y_i=y_j}\}|}{|\mathcal{E}|} \\
    h_{node} &= \frac{1}{|\mathcal{V}|}\sum_{v_i\in \mathcal{V}}\frac{|\{{v_{j}|v_{j}\in \mathcal{N}_i, y_i=y_j}\}}{d_{i}}
\end{split}
\end{equation}
where the edge homophily $h_{edge}$ represents the ratio of homophilous edges in the entire graph and the node homophily $h_{node}$ calculates the average homophily degree across all nodes. Since $h_{edge}$ and $h_{node}$ show little difference in the datasets used in this paper, we only consider and discuss edge homophily, denoted as $h$, in the following section.
\section{Methodology}
In this section, we first introduce the definition of heterophilous information in Sect. \ref{sec:hetero_info}. Next, we analyze how heterophilous information improves graph learning in Sect. \ref{sec:ana_hi}. Finally, we introduce our proposed method \model{} in Sect. \ref{sec:hignn}.

\subsection{Heterophilous Information}\label{sec:hetero_info}
In heterophilous graphs, nodes with different labels tend to be connected. Consequently, many studies\cite{GloGNN,CPGNN,WRGAT} perceive heterophily as a challenge in graph representation learning and propose methods to address the heterophily by utilizing node features or structural information. During the process, however, the heterophilous information, which could be helpful for graph representation learning, is discarded. Next, we give the definition of heterophilous information and discuss its benefits in improving graph learning.

\begin{definition}
For any node $u$ in the graph $\mathcal{G}$, the heterophilous information $\mathcal{H}$ is defined as its neighbor distribution, which describes how nodes become heterophilous:
\begin{equation}
    \mathcal{H}_u = [p_1, p_2, \cdots, p_c], where\ p_i = \frac{|\{ v|v\in \mathcal{N}_u, y_v=i \}|}{|\mathcal{N}_u|}
\label{eq:hetero_info}
\end{equation}
\label{def:hetero_info}
\end{definition}

Here each element $p_i$ at position $i$ in $\mathcal{H}_u$ describes the probability of node neighbors $\mathcal{N}_u$ belongs to class $i$. Compared with the definition of node homophily in Eq. (\ref{equ:def_hom}), heterophilous information better characterizes the heterophilous context of a node: for any node $u$ with class $i$, when its homophily is very high e.g. $h\to 0$, the $\mathcal{H}_u$ is approximate to an all-zero vector with $p_i\to 1$. On the contrary, low homophily results in the $p_i\to 0$ in $\mathcal{H}_u$. In this case, the other entries of $\mathcal{H}_u$ describe how the node becomes heterophilous, while $h$, a single scalar value, is insufficient to describe the heterophilous contexts of a node.

Next, to utilize the heterophilous information, we create a new adjacency matrix $A'$ by connecting the nodes that share a high similarity of $\mathcal{H}$, where Each entry $A'_{i,j}$ can be expressed as
\begin{equation}
    A'_{i,j} = \mathbbm{1}(cos(\mathcal{H}_i,\mathcal{H}_j)>\delta)
\label{eq:hi_adj}
\end{equation}
where $\mathbbm{1}$ is an indicator function that is equal to $1$ when the cosine similarity of heterophilous information of nodes $i$ and $j$ larger than a threshold $\delta$, otherwise, it is equal to $0$.

\subsection{Analysis of Heterophilous Information}\label{sec:ana_hi}
After defining the new adjacency matrix $A'$ with heterophilous information, we analyze how well this approach improves graph learning by theoretically examining the homophily degree $\hat{h}$ of $A'$. Therefore, we first make the following assumptions: 
\begin{itemize}
    \item Every node has a probability of $h$ to connect with intra-class nodes, and a probability of $\frac{1-h}{c-1}$ to connect with inter-class nodes. Each probability is corrupted by an independent Gaussian noise $\epsilon \sim N(0,\sigma^2)$.
    \item The number of nodes is balanced across each class.
\end{itemize}

Then we have the following theorem under the assumptions.

\begin{theorem}
The homophily degree $\hat{h}$ of $A'$ is directly proportional to the threshold $\delta$, inversely proportional to the variance $\sigma^2$ of sampling noise, and directly proportional to the distance between the original homophily degree and reciprocal of the class number $c$:
\label{theo:new_hom}
\begin{equation}
\hat{h}\propto \delta,\ \hat{h}\propto \frac{1}{\sigma^2},\ \hat{h}\propto \lvert h-\frac{1}{c} \rvert
\end{equation}
\end{theorem}

$Proof.$ Following the aforementioned assumptions, we ensure the original homophily degree is $h$. Then we can get the heterophilous information of node $u$ as a vector:
\begin{equation}
    \mathcal{H}_u = [p_1, p_2, \cdots, p_c],
    \text{where}\ p_i = 
      \begin{cases}
        h+\epsilon_{u,i} & \text{if } i=y_u \\
        \frac{1-h}{c-1}+\epsilon_{u,i} & \text{if } i\neq y_u
      \end{cases}
\end{equation}

Then we define the similarity of the neighbor distribution of node $u$ and $v$ as $\b{S^+}$ when $y_u=y_v$, and $\b{S^-}$ when $y_u\neq y_v$. Then we have

\begin{equation}
    \begin{split}
        \b{S^+}&=\cos(\mathcal{H}_u,\mathcal{H}_v)\\
        &=\frac{1}{\mathcal{H}_{norm}}\biggl( (h+\epsilon_{v,y_v})(h+\epsilon_{u,y_u})\\
        &+\sum_{\substack{i=1,\\ i\neq y_u}}^c (\frac{1-h}{c-1}+\epsilon_{u,i})(\frac{1-h}{c-1}+\epsilon_{v,i}) \biggl)\\
        \b{S^-}&=\cos(\mathcal{H}_u,\mathcal{H}_v) \\
        &=\frac{1}{\mathcal{H}_{norm}} \biggl( (h+\epsilon_{u,y_u})(\frac{1-h}{c-1}+\epsilon_{v,y_v})\\
        &+(h+\epsilon_{v,y_v})(\frac{1-h}{c-1}+\epsilon_{u,y_u}) \biggl. \\
        \biggl. &+\sum_{\substack{i=1,\\ i\neq y_u, i\neq y_v}}^c (\frac{1-h}{c-1}+\epsilon_{u,i})(\frac{1-h}{c-1}+\epsilon_{v,i}) \biggl)
    \end{split}
\end{equation}

where $\mathcal{H}_{norm}=\norm{\mathcal{H}_u}\norm{\mathcal{H}_v}$.

The new homophily degree $\hat{h}$ of the newly constructed $A'$ can be expressed as the ratio of homophilous edges to the total number of edges
\begin{equation}
    \hat{h} = \frac{P(\b{S^+}>\delta)}{P(\b{S^+}>\delta)+(c-1)P(\b{S^-}>\delta)}
    \label{eq:new_hom}
\end{equation}

To simplify the analysis, we ignore the second order of noise term $\epsilon_{u,i}\epsilon_{v,i}$ since its magnitude is much smaller than the other terms. We also treat the normalization term $\mathcal{H}_{norm}$ as a constant, by omitting the noise term, as $\mathcal{H}_{norm}$ is identical for both the $\b{S^+}$ and $\b{S^-}$. Consequently, we can express $\b{S^+}$ as
\begin{equation}
    \begin{split}
        \b{S^+} = \mathcal{H}_{norm}^{-1}\biggl(h^2+\frac{(1-h)^2}{c-1}+\epsilon^+\biggl)
    \end{split}
\end{equation}
where 
\begin{equation}
    \begin{split}
        \epsilon^+ &= h(\epsilon_{u,y_u}+\epsilon_{v,y_v})+\sum_{\substack{i=1, i\neq y_u}}^c \frac{1-h}{c-1}(\epsilon_{u,i}+\epsilon_{v,i}) \\
        &\sim N(0,(h^2+\frac{(1-h)^2}{c-1})2\sigma^2)
    \end{split}
\end{equation}

Then we can get the probability of homophilous edges as
\begin{equation}
    \begin{split}
        P(\b{S^+}>\delta) &= P\biggl(\Bigl(\mathcal{H}_{norm}^{-1}(h^2+\frac{(1-h)^2}{c-1}+\epsilon^+)\Bigl) >\delta\bigg)\\
        &= P\biggl(\epsilon^+> \delta \mathcal{H}_{norm}-h^2-\frac{(1-h)^2}{c-1} \bigg)\\
        &=\Phi\left(\frac{h^2+\frac{(1-h)^2}{c-1}-\delta \mathcal{H}_{norm}}{(h^2+\frac{(1-h)^2}{c-1})^{\frac{1}{2}}\sqrt{2}\sigma}\right)
    \end{split}
\end{equation}

Let
\begin{equation}
    t_+ = \frac{h^2+\frac{(1-h)^2}{c-1}-\delta \mathcal{H}_{norm}}{(h^2+\frac{(1-h)^2}{c-1})^{\frac{1}{2}}\sqrt{2}\sigma}
\end{equation}

we have $P(\b{S^+}>\delta)=\Phi(t_+)$. 

Similarly, for $\b{S^-}$, we follow the same process as $\b{S^+}$ and we can get $P(\b{S^-}>\delta) = \Phi(t_-)$, where
\begin{equation}
    t_- = \frac{2h\frac{1-h}{c-1}+(C-2)(\frac{1-h}{c-1})^2-\delta \mathcal{H}_{norm}}{(h^2+\frac{(1-h)^2}{c-1})^{\frac{1}{2}}\sqrt{2}\sigma}
\end{equation}

Next, we substitude $P(\b{S^+}>\delta)$ and $P(\b{S^-}>\delta)$ into Eq. \ref{eq:new_hom} to get the new homophily degree
\begin{equation}\label{eq:hat_hom}
    \begin{split}
        \hat{h} = \left( 1+(c-1)\frac{\Phi(t_-)}{\Phi(t_+)} \right)^{-1}
    \end{split}
\end{equation}

We then analyze the influence of the threshold $\delta$, the standard variance of noise $\sigma$, and the original homophily degree $h$ by taking partial derivative with respect to $\hat{h}$
\begin{equation}
    \begin{split}
        \frac{\partial \hat{h}}{\partial \delta}\geq 0, \; \frac{\partial \hat{h}}{\partial \sigma}\leq 0,\; 
    \frac{\partial \hat{h}}{\partial h}
        \begin{cases}
        \geq 0 & \text{if } h \geq \frac{1}{c} \\
        < 0 & \text{if } h < \frac{1}{c}
        \end{cases}
    \end{split}
    \label{eq:partial}
\end{equation}

From the partial derivative results, we can get Theorem \ref{theo:new_hom} proved.

\begin{figure*}[t]
    \centering
    \subfloat[
    ]{%
       \includegraphics[width=0.33\textwidth]{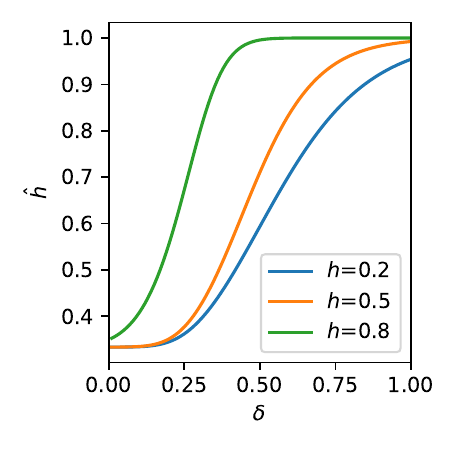}\label{fig:proof_delta}}
    \hfill
    \subfloat[
    ]{%
        \includegraphics[width=0.33\textwidth]{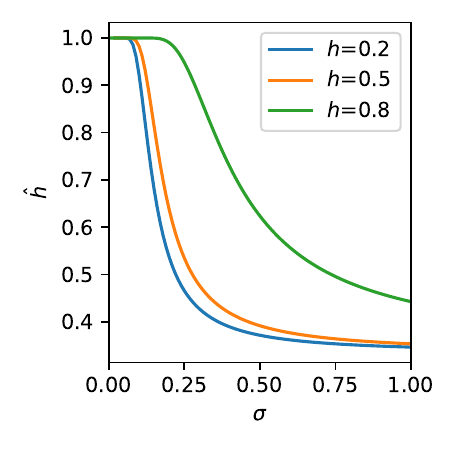}\label{fig:proof_sigma}}
    \hfill
    \subfloat[
    ]{%
        \includegraphics[width=0.33\textwidth]{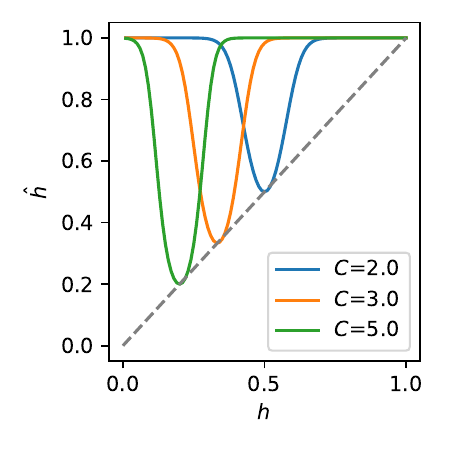}\label{fig:proof_Cls}}
    \hfill
    \caption{Relation of improved homophily and original homophily under different parameter setting. Fig. (a), (b), and (c) shows the relation under different $\delta$, $\sigma$, and $c$ respectively.}
    \label{fig:proof}
\end{figure*}

To better illustrate how the improved homophily $\hat{h}$ is affected, we plot the influences of these parameters according to the theoretical results in Eq. (\ref{eq:hat_hom}). As depicted in Fig. \ref{fig:proof_delta} and Fig. \ref{fig:proof_sigma}, $\hat{h}$ is directly proportional to $\delta$ and inversely proportional to $\sigma$ under different levels of $h$. These results align with our intuition: a larger noise during neighbor sampling makes it more challenging to capture their true neighbor distribution, thereby reducing $\hat{h}$. Moreover, a higher $\delta$ allows fewer edges with a higher similarity of neighbor distribution to connect, thereby increasing $\hat{h}$. For the influence of $h$, as shown in Fig. \ref{fig:proof_Cls}, we can see that a larger distance between $h$ and $\frac{1}{c}$ improves $\hat{h}$. When $h=\frac{1}{c}$, the probability of edges connecting to intra-class or inter-class nodes becomes identical i.e. $P(\b{S^+}>\delta)=P(\b{S^-}>\delta)$, which implies that there is no useful information from neighbors because the aggregated messages from neighbors become indistinguishable for each class. Conversely, if $h$ is very high or low, the neighbor distribution becomes distinguishable for each class, leading to an improved $\hat{h}$. This phenomenon has also been observed in the theoretical analysis by \cite{when_do_graph_help}, which is known as the mid-homophily pitfall. 

We then measure the influences of these parameters on real-world datasets. As depicted in Table \ref{tab:hom_improv_hetero}, we measure the average standard deviation $\bar{\sigma}$, original homophily $h$, and the homophily improvement $\hat{h}-h$ under different threshold $\delta$ across 6 datasets. The results of $\hat{h}-h$ are highlighted in \textbf{bold} if there is a positive improvement. The results align with our theoretical analysis: a larger $\delta$ improves $\hat{h}$ while a larger $\bar{\sigma}$ decreases $\hat{h}$. It is worth noting that the $\bar{\sigma}$ is typically small, resulting in a large improvement in $\hat{h}$ as shown in \ref{fig:proof_delta}. This shows the newly constructed $A'$ effectively improves the graph learning. This is further evidenced in Table \ref{tab:hom_improv_hetero} where most datasets exhibit a significant increase of homophily with a high $\sigma$. This observation validates the efficacy of leveraging heterophilous information to improve the performance of GNNs.

\begin{table}[!t]
    \centering\caption{Homophily Improvement on Heterophilous Datasets}
    \resizebox{1\hsize}{!}{
    \begin{tabular}{c *{6}{c}}
    \toprule
        Dataset & Texas & Squirrel & Wisconsin & Chameleon & Cornell & Actor \\ \midrule
$\bar{\sigma}$& 0.2328& 0.2623& 0.2837& 0.2752& 0.2471& 0.3122\\
$h$ & 0.0609 & 0.1778 & 0.2960 & 0.2299 & 0.2221 & 0.2167\\ \midrule
~~$\hat{h}-h$, $\delta=0.3$~~ & \textbf{ 0.4189} & \textbf{ 0.1768} & \textbf{ 0.0863} & -0.0133 & -0.0233 & -0.0034\\
~~$\hat{h}-h$, $\delta=0.6$~~ & \textbf{ 0.4634} & \textbf{ 0.2357} & \textbf{ 0.0940} & -0.0005 & -0.0206 & -0.0037\\
~~$\hat{h}-h$, $\delta=0.8$~~ & \textbf{ 0.4827} & \textbf{ 0.3399} & \textbf{ 0.0951} & \textbf{ 0.0212} & -0.0142 & -0.0037\\
~~$\hat{h}-h$, $\delta=0.9$~~ & \textbf{ 0.5404} & \textbf{ 0.4238} & \textbf{ 0.0865} & \textbf{ 0.0605} & -0.0051 & -0.0040\\
~~$\hat{h}-h$, $\delta=1.0$~~ & \textbf{ 0.6447} & \textbf{ 0.5152} & \textbf{ 0.1129} & \textbf{ 0.2448} & \textbf{ 0.1743} & -0.0030\\ \bottomrule
    \end{tabular}
    }
    \label{tab:hom_improv_hetero}
\end{table}



\subsection{HiGNN}\label{sec:hignn}
In this subsection, we present our method, Heterophilous Information-enhanced Graph Neural Networks(HiGNN), which is based on the preceding analysis of heterophilous information.

\begin{figure*}
    \centering
    \includegraphics[width=0.7\textwidth]{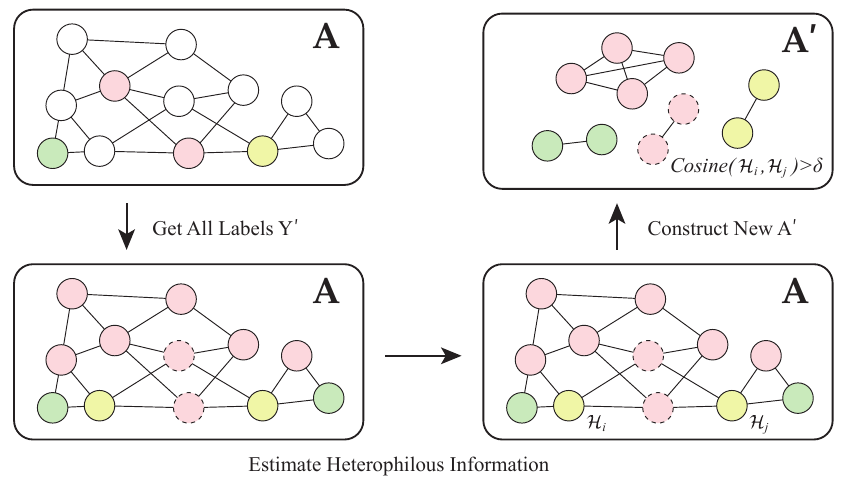}
    \caption{The construction process of new adjacency matrix $A'$ with heterophilous information. We first obtain all labels $Y'$ by training an off-the-shelf model on the training data. Then based on the estimated labels, we measure the heterophilous information $\mathcal{H}$ on all the nodes. Finally, nodes with similar $\mathcal{H}$ are connected in the $A'$.}
    \label{fig:new_A}
\end{figure*}

\textbf{Graph structure construction.} Figure. \ref{fig:new_A} illustrates the process of the construction of a new adjacency matrix $A'$ by incorporating heterophilous information $\mathcal{H}$. As shown on the left side, we first obtain all the labels $Y'$. Here we follow the conventional setting of graph-based semi-supervised learning\cite{pesudo_label1,pesudo_label2} to estimate the labels $A'$ of all nodes by training an off-the-shelf model on the training data $X_{train}$ and $Y_{train}$:
\begin{equation}\label{equ:all_pred}
    \min_{\mathcal{M}} \mathcal{L}( \mathcal{M}(X_{train}),Y_{train}),\ 
    Y' = \mathcal{M}(X)
\end{equation}
where $\mathcal{M(\cdot)}$ represents an off-the-shelf model and $\mathcal{L}$ denotes its loss function. After obtaining all the estimated labels $Y'$, we derive the heterophilous information $\mathcal{H}$ of all nodes by their neighbor distribution using Eq. (\ref{eq:hetero_info}). Finally, based on $\mathcal{H}$, we construct $A'$ by preserving connections with high similarity of $\mathcal{H}$ using Eq. (\ref{eq:hi_adj}).

For the nodes exhibiting high homophily or heterophily, denoted as red nodes or yellow nodes respectively in Fig. \ref{fig:new_A}, our method effectively groups them within the same class. Additionally, our method can also connect nodes with half-homophily, represented by a dashed stroke in Fig. \ref{fig:new_A}. These half-homophily nodes, which act as boundary nodes straddling two classes, share similar semantics. While connecting boundary nodes with nodes inside the class boundary increases homophily, it may have detrimental effects on graph learning. This is due to the influence of connections with other types of nodes on boundary nodes, leading to different patterns that manifest as mediators. Our new adjacency matrix establishes connections among boundary nodes and facilitates message passing among nodes with similar semantics, thereby contributing to effective graph learning at various homophily levels.

Although this label estimation is not flawless and may result in a biased estimation of heterophilous information $\mathcal{H}$, the $\mathcal{H}$ defined in this paper is error-tolerant, accommodating differences between neighbor distributions in similar nodes. As illustrated in \ref{fig:proof_delta}, for $\delta$ that is close to $1$, the improved homophily $\hat{h}$ does not decrease significantly. Our experimental analysis of homophily improvement across 9 real-world datasets in Fig. \ref{fig:hom_degree} further substantiates this proposition. 

\textbf{Channel Fusion.} Although the newly constructed adjacency matrix $A'$ contains rich heterophilous information with improved homophily degree, the original adjacency matrix $A$ is informative as well. To retain the original graph topological information as a supplement, we employ a late fusion strategy on these two graph topologies. Specifically, for the node embeddings $Z_{l-1}$ at layer $l-1$, we perform graph convolution on $A'$ and $A$ to obtain the updated node embeddings of the next layer $Z^l_{new}$ and $Z^l_{old}$ respectively,
\begin{equation}
    Z^{l}_{new} = g(\hat{A}'Z^{l-1}W^{l-1}),Z^{l}_{old} = g(\hat{A}Z^{l-1}W^{l-1})
\end{equation}
where $\hat{A}'$ and $\hat{A}$ denote the normalized adjacency matrices of $A'$ and $A$ respectively, $g(\cdot)$ is an activation function, and $W^{l-1}$ is a weight matrix. Given that the heterophilous information is a new addition and its contribution to the final result could vary depending on specific datasets, we introduce a balance parameter $\lambda$ for the late fusion of the embeddings from two channels, which could be expressed as:
\begin{equation}
    Z^{l} = \lambda Z^{l}_{new} + Z^{l}_{old}
\end{equation}

The balance parameter $\lambda$ is used to regulate the proportion of heterophilous information incorporated in the learning process. A higher $\lambda$ emphasizes heterophilous information, while a lower $\lambda$ emphasizes original topological information. The choice of late fusion, as opposed to combining the two adjacency matrices early, was deliberately taken to keep the heterophilous and topological information unperturbed from each other. This approach ensures that the unique contributions of each can be captured separately, allowing each to fully express the nuances of its information channels.

\subsection{Complexity Analysis}
The time complexity of \model{} mainly comes from three parts: (1) label estimation with GNNs $\mathcal{O}(L\lvert \mathcal{V} \rvert MF+L\lvert \mathcal{E} \rvert F)$, (2) heterophilous information collection $\mathcal{O}(\lvert \mathcal{E} \rvert)$, (3) new adjacency matrix construction $\mathcal{O}(c\lvert \mathcal{V} \rvert^2)$, and (4) graph convolution on two channels $\mathcal{O}(2L\lvert \mathcal{V} \rvert MF+2L\lvert \mathcal{E} \rvert F)$, where $L$, $M$, $F$, and $c$ refer to the number of graph convolution layers, node feature dimension, hidden embedding size, and class number respectively. The complexity of graph convolution is linear to the general GNNs baselines and the total complexity of constructing new adjacency matrix $A'$ with heterophilous information is $\mathcal{O}(L\lvert \mathcal{V} \rvert MF+L\lvert \mathcal{E} \rvert F+\lvert \mathcal{E} \rvert+c\lvert \mathcal{V} \rvert^2)$, which is comparable to the other neighbor extension methods.
\section{Experiments}

\subsection{Experimental Setup}
\textbf{Baselines}. We benchmark \model{} against a variety of methods, including non-graph methods, general GNNs, and heterophily-oriented GNNs. Specifically, we use MLP as the graph-agnostic method and GCN\cite{GCN}, GAT\cite{GAT}, SGC\cite{sgc}, Mixhop\cite{Mixhop}, and GraphSage\cite{graphSage} as the general GNNs baselines. For the heterophily-oriented GNNs, we use GCN2\cite{GCN2}, H2GCN\cite{H2GCN}, LINKX\cite{LINKX},  ACMGCN\cite{ACM-GCN}, and GPRGNN\cite{GPR-GNN}, GloGNN\cite{GloGNN}, WRGAT\cite{WRGAT}, and GGCN\cite{GGCN}.

\textbf{Datasets.} We measure all the methods on 3 homophilous datasets\footnote{\url{https://github.com/kimiyoung/planetoid/tree/master/data}} and 6 heterophilous graph datasets\footnote{\url{https://github.com/CUAI/Non-Homophily-Large-Scale/tree/master/data}}. The statistics of all the datasets are presented in Table \ref{tab:graph_statistics}, where the homophilous graphs including Cora, CiteSeer, and PubMed, exhibit a significantly higher homophily compared to heterophilous graphs. We run 10 times for each dataset following public random splits with the ratio of 60\%/20\%/20\% for train, validation, and test sets respectively\footnote{\url{https://github.com/bingzhewei/geom-gcn/tree/master/splits}}.

\begin{table*}[!ht]
    \centering\caption{Dataset Statistics}
    \resizebox{0.9\hsize}{!}{
    \begin{tabular}{l|ccccccccc}
    \toprule
        Dataset & Cora & CiteSeer & PubMed & Chameleon & Cornell & Actor & Squirrel & Texas & Wisconsin  \\ \midrule
        \#Nodes & 2,708 & 3,327 & 19,717 & 2,277 & 183 & 7,600 & 5,201 & 183 & 251 \\
        \#Edges & 10,556 & 9,104 & 88,648 & 38,328 & 478 & 37,526 & 222,134 & 492 & 750 \\
        \#Features & 1,433 & 3,703 & 500 & 2,325 & 1,703 & 932 & 2,089 & 1,703 & 1,703 \\
        \#Classes & 7 & 6 & 3 & 5 & 5 & 5 & 5 & 5 & 5 \\
        $h_{edge}$ & 0.81 & 0.74 & 0.80 & 0.23 & 0.30 & 0.22 & 0.22 & 0.06 & 0.17 \\
        $h_{node}$ & 0.83 & 0.72 & 0.79 & 0.25 & 0.39 & 0.22 & 0.22 & 0.10 & 0.15 \\ \toprule
    \end{tabular}}
    \label{tab:graph_statistics}
\end{table*}

\textbf{Implementations.}
To generate label predictions for neighbor distribution, we select ACM-GCN as an off-the-shelf model. Given the flexibility of our method, which can be integrated into any GNN method, we adopt GCN as the backbone model, enhanced with a high-pass filter\cite{ACM-GCN}. Then we run \model{} and all the baselines on 9 datasets, implemented using Pytorch on an A5000 GPU. The models are trained until convergence is achieved, with an early stopping criterion of 40 epochs or a maximum of 2000 epochs. For all the methods, we search for the optimal hyper-parameters, including hidden embedding size from $\{64, 128, 256\}$, learning rate from $\{10^{-2}, 10^{-3}\}$, weight decay from $\{5\times 10^{-4}, 5\times 10^{-5}\}$, number of layers from $\{1, 2\}$ and dropout rate from $\{0.3, 0.4, 0.5, 0.6, 0.7\}$ on the validation sets for each dataset. Specifically for \model{}, we search for the threshold $\delta$ from $\{0.5, 0.8, 0.9, 1.0\}$ and lambda $\lambda$ from $\{1, 0.5, 10^{-1}, 5\times 10^{-2}, 1\times 10^{-2}, 1\times 10^{-3}, 1\times 10^{-4}\}$ to control the amount of heterophilous information involved. The final hyper-parameters for HiGNN are provided in Table \ref{tab:hyperpara_hignn}.

\subsection{Performance}

Table \ref{tab:nc_acc} shows a comparison of \model{} with other baselines. All methods are evaluated with node classification tasks, where both the mean accuracy and standard deviation are reported based on 10 public random splits. The average rank for each method is calculated across 9 datasets, with a lower rank indicating better performance. For each dataset, the best-performing methods are highlighted in \textbf{bold} and the runner-up methods are \underline{underlined}. For methods that augment general GNN baselines with heterophilous information (denoted as +Hi), we halve the embedding space to control the total number of parameters and time complexity, ensuring a fair comparison. We use OOM to denote out-of-memory errors

First, our proposed \model{} generally outperforms all the other methods, achieving the lowest average rank, indicating the effectiveness of incorporating  heterophilious information into graph learning. \model{} enhances the performance on both the homophilous and heterophilous datasets, corroborating the explanation provided in Sect. \ref{sec:ana_hi} that the new matrix constructed by heterophilous information improves graph learning at different homophily levels. Besides, the effectiveness of heterophilous information is also shown in the improvement on the general GNNs baselines where the performance is improved on most of the datasets. Second, compared with general GNN baselines such as GCN and GAT, MLP performs poorly on homophilous graphs but still demonstrates strong performance on heterophilous graphs. This validates the assertion that heterophily impedes graph convolution in graph learning. Then for the methods that aggregate embeddings of k-hop neighbors during the message passing, Mixhop, H2GCN, GGCN, CPGCN, and GPR-GNN show their improvement with respect to GCN, but they still have a large gap with \model{} and GloGNN that exploit global similar nodes. This indicates that for heterophilous graphs, focusing on k-hop neighbors is not as sufficient as constructing new graph structures that capture neighbors with similar semantics. Note that although GloGNN delivers the best results on 3 datasets, it requires the calibrated adjustment of many parameters\footnote{
Noted that some baseline results presented in Table3 maybe different from those in the original papers on some datasets. The performance of GloGNN\cite{GloGNN} reported there is different as in their original paper because we use the preprocessed heterophilous datasets from \cite{lim2021large}}. In contrast, our method is more straightforward and flexible, requiring fewer parameter tunings.

\begin{table*}[!ht]
    \centering\caption{Node classification performance of \model{}}
    \resizebox{\textwidth}{!}{
    \begin{tabular}{lcccccccccc}
    \toprule
        Methods & Cora & CiteSeer & PubMed & Chameleon & Cornell & Actor & Squirrel & Texas & Wisconsin & Rank$\downarrow$  \\ \midrule
GCN& 86.90 $\pm$ 1.09& 75.33 $\pm$ 2.07& 87.85 $\pm$ 0.37& 49.21 $\pm$ 2.56& 63.24 $\pm$ 6.40& 31.08 $\pm$ 0.65& 26.57 $\pm$ 1.47& 57.57 $\pm$ 4.24& 57.65 $\pm$ 5.33& 15.17 \\
GCN+Hi& {87.04 $\pm$ 1.24}& {75.73 $\pm$ 2.26}& {88.40 $\pm$ 0.57}& {61.18 $\pm$ 2.51}& {65.95 $\pm$ 6.00}& {31.74 $\pm$ 0.97}& {28.99 $\pm$ 3.44}& {60.00 $\pm$ 3.99}& {59.61 $\pm$ 7.58}& 12.44 \\ \midrule
GAT& 86.06 $\pm$ 1.31& 74.65 $\pm$ 1.26& 86.75 $\pm$ 0.58& 68.46 $\pm$ 1.99& 56.22 $\pm$ 5.06& 27.84 $\pm$ 1.46& \underline{60.46 $\pm$ 2.18} & 57.57 $\pm$ 8.92& 54.90 $\pm$ 6.85& 13.72 \\
GAT+Hi& {86.78 $\pm$ 1.80}& {75.32 $\pm$ 1.63}& {87.08 $\pm$ 0.47}& \underline{{68.49 $\pm$ 1.66}} & {58.38 $\pm$ 4.80}& {28.14 $\pm$ 1.12}& \textbf{{61.37 $\pm$ 2.10}} & 57.03 $\pm$ 7.59& {58.04 $\pm$ 6.87}& 11.89 \\ \midrule
GraphSage& 80.44 $\pm$ 1.90& 74.53 $\pm$ 1.62& 86.54 $\pm$ 0.48& 51.34 $\pm$ 2.02& 74.59 $\pm$ 5.58& 34.12 $\pm$ 1.79& 31.55 $\pm$ 2.26& 70.81 $\pm$ 9.35& 79.41 $\pm$ 3.61& 13.22 \\
GraphSage+Hi& {80.78 $\pm$ 2.03}& {74.98 $\pm$ 1.21}& {87.14 $\pm$ 0.47}& {52.43 $\pm$ 2.34}& 73.24 $\pm$ 6.17& 34.03 $\pm$ 1.17& {32.05 $\pm$ 1.79}& 70.27 $\pm$ 8.73& {80.20 $\pm$ 3.97}& 12.11 \\ \midrule
SGC& 84.97 $\pm$ 1.95& 75.66 $\pm$ 1.37& 87.15 $\pm$ 0.47& 64.78 $\pm$ 1.98& 55.41 $\pm$ 5.29& 25.83 $\pm$ 1.09& 41.78 $\pm$ 2.88& 58.11 $\pm$ 6.27& 57.84 $\pm$ 4.83& 14.11 \\
SGC+Hi& {85.25 $\pm$ 1.58}& {76.13 $\pm$ 2.26}& {88.84 $\pm$ 0.73}& 61.86 $\pm$ 2.92& {71.89 $\pm$ 6.27}& {34.17 $\pm$ 0.95}& {41.98 $\pm$ 1.16}& {64.05 $\pm$ 7.21}& {73.53 $\pm$ 5.79}& 10.00 \\ \midrule
Mixhop& 86.58 $\pm$ 1.12& 75.14 $\pm$ 1.62& 88.53 $\pm$ 0.48& 62.35 $\pm$ 1.48& 65.14 $\pm$ 6.91& 31.76 $\pm$ 2.06& 40.16 $\pm$ 5.78& 60.81 $\pm$ 6.40& 66.86 $\pm$ 5.18& 12.33 \\
Mixhop+Hi& {87.20 $\pm$ 0.96}& {75.72 $\pm$ 1.59}& {88.90 $\pm$ 0.55}& {63.31 $\pm$ 2.47}& {66.22 $\pm$ 3.87}& {33.70 $\pm$ 1.26}& {42.47 $\pm$ 2.19}& {62.70 $\pm$ 6.22}& {69.02 $\pm$ 5.76}& 9.67 \\ \midrule \midrule
MLP& 74.25 $\pm$ 2.16& 72.09 $\pm$ 1.34& 87.07 $\pm$ 0.30& 49.87 $\pm$ 2.41& 77.57 $\pm$ 7.43& 35.65 $\pm$ 0.66& 32.80 $\pm$ 1.43& 76.22 $\pm$ 8.24& 81.76 $\pm$ 4.98& 11.61 \\ \midrule
ACM-GCN& 87.83 $\pm$ 0.95& 75.56 $\pm$ 1.32& \underline{89.48 $\pm$ 0.58} & 67.94 $\pm$ 1.68& 77.57 $\pm$ 5.26& 35.09 $\pm$ 1.18& 53.35 $\pm$ 1.33& \underline{82.70 $\pm$ 6.27} & \underline{83.53 $\pm$ 3.83} & 4.39 \\ \midrule
GCN2& 86.80 $\pm$ 1.08& 74.84 $\pm$ 1.48& 88.27 $\pm$ 0.72& 65.07 $\pm$ 2.71& 54.32 $\pm$ 9.14& 33.48 $\pm$ 2.05& 52.68 $\pm$ 0.98& 61.89 $\pm$ 6.43& 56.86 $\pm$ 8.32& 12.56 \\ \midrule
H2GCN& 87.71 $\pm$ 1.25& 76.32 $\pm$ 1.54& 89.17 $\pm$ 0.45& 65.88 $\pm$ 2.38& 75.95 $\pm$ 7.37& 36.23 $\pm$ 0.98& 57.08 $\pm$ 1.58& 75.68 $\pm$ 6.74& 80.59 $\pm$ 2.99& 4.78 \\ \midrule
GPR-GNN& 87.42 $\pm$ 1.21& 75.43 $\pm$ 1.47& 89.18 $\pm$ 0.51& 67.17 $\pm$ 1.47& 74.32 $\pm$ 3.66& 35.47 $\pm$ 1.66& 43.84 $\pm$ 3.07& 74.86 $\pm$ 5.70& 79.61 $\pm$ 5.56& 6.78 \\ \midrule
WRGAT& 75.47 $\pm$ 2.90& 74.14 $\pm$ 1.40& OOM& 51.40 $\pm$ 2.24& 76.49 $\pm$ 6.75& 36.15 $\pm$ 1.00& 30.73 $\pm$ 1.70& 76.76 $\pm$ 4.07& 79.61 $\pm$ 5.16& 12.22 \\ \midrule
GGCN& 86.32 $\pm$ 0.91& 76.65 $\pm$ 1.91& 88.25 $\pm$ 0.43& 56.49 $\pm$ 2.65& 71.35 $\pm$ 7.34& 34.86 $\pm$ 0.87& 39.18 $\pm$ 1.87& 65.14 $\pm$ 8.30& 74.12 $\pm$ 5.37& 10.33 \\ \midrule
LINKX& 77.32 $\pm$ 1.68& 72.00 $\pm$ 1.90& 78.39 $\pm$ 1.09& 68.38 $\pm$ 2.50& 39.46 $\pm$ 17.89& 27.06 $\pm$ 1.22& 59.15 $\pm$ 2.06& 52.43 $\pm$ 9.21& 55.88 $\pm$ 6.29& 15.78 \\ \midrule
GloGNN& \underline{88.31 $\pm$ 1.15} & \underline{77.41 $\pm$ 1.65} & \textbf{89.62 $\pm$ 0.35} & 65.59 $\pm$ 2.21& \textbf{82.16 $\pm$ 5.82} & \textbf{37.36 $\pm$ 1.34} & 29.44 $\pm$ 1.36& 69.19 $\pm$ 11.16& 82.35 $\pm$ 5.11& 5.00 \\ \midrule
HiGNN& \textbf{89.72 $\pm$ 1.46} & \textbf{79.30 $\pm$ 2.13} & 89.43 $\pm$ 0.53& \textbf{68.86 $\pm$ 1.45} & \underline{80.00 $\pm$ 4.26} & \underline{37.21 $\pm$ 1.35} & 54.78 $\pm$ 1.58& \textbf{86.22 $\pm$ 4.67} & \textbf{85.88 $\pm$ 3.18} & 1.89 \\ \midrule
    \end{tabular}}
    \label{tab:nc_acc}
\end{table*}

\subsection{Improvement in Homophily Degree}

\begin{figure}[h]
    \centering
    \includegraphics[width=0.5\textwidth]{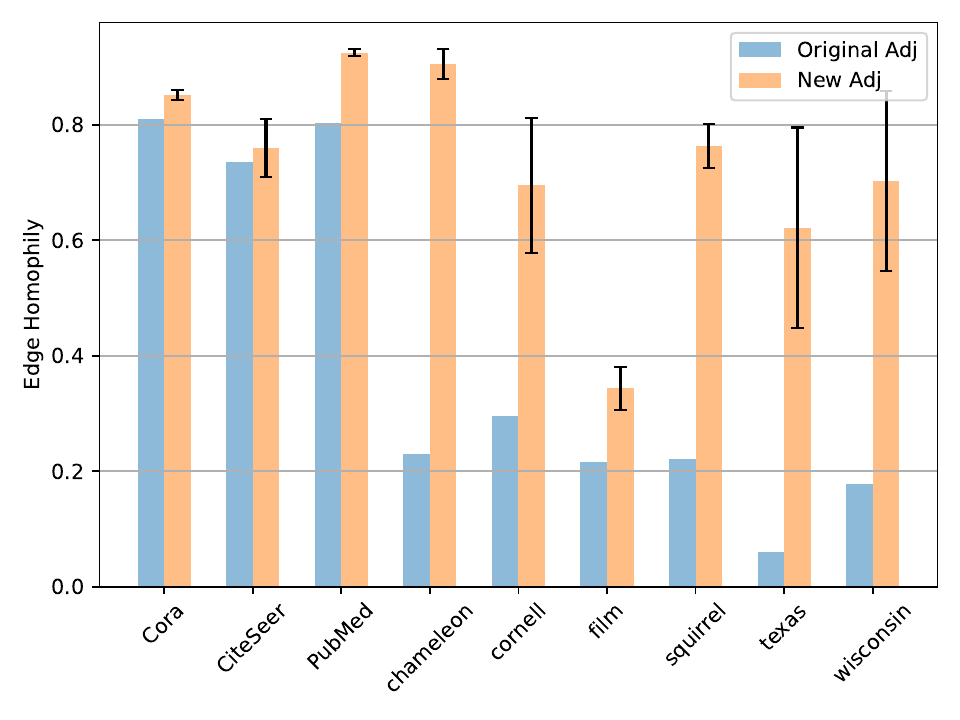}
    \caption{The homophily improvement after constructing new adjacency matrices with heterophilous information. For each dataset, we show the mean and standard deviation of the newly constructed matrix based on 10 splits.}
    \label{fig:hom_degree}
\end{figure}

To validate that the new graph structure $A'$ with heterophilous information in \model{} improves homophily, we measure the homophily degree for both the original graph structure $A$ and the new graph structure $A'$. Specifically, we compute homophily degrees across 9 datasets on 10 random splits and then report the mean and standard deviation. As depicted in Fig. \ref{fig:hom_degree}, the newly constructed matrix improves the homophily degree on all datasets, particularly for heterophilous datasets. This suggests that heterophilous information is helpful in grouping semantically similar nodes. It is worth noting that the estimation of heterophilous information is based on the predicted labels instead of the true labels, demonstrating the estimation is error-tolerant as suggested in Sect. \ref{sec:hignn}.

\subsection{Ablation Study}

To demonstrate the performance improvement of \model{} achieved through the incorporation of heterophilous information and the efficacy of the fusion strategy employed in \model{}, we undertake an ablation study. As shown in Table \ref{tab:abla}, for \model{}, we examine the influence of channels by removing the new channel (w/o $A'$), removing the original channel (w/o $A$), or fusing two adjacency matrices $A$ and $A'$ in one channel ($A$+$A'$). The results show that the performance of \model{} drops when either the new channel or the original channel is removed. This suggests that optimal performance is attained by considering different aspects of these two channels. Furthermore, the early fusion of two adjacency matrices ($A$+$A'$) underperforms compared to the late fusion in \model{}, indicating the late fusion strategy allows the model to fully capture the unique contributions of each channel as suggested in Sect. \ref{sec:hignn}.

\begin{table*}
    \centering\caption{Ablation study on \model{}}
    \resizebox{\textwidth}{!}{
    \begin{tabular}{lcccccccccc}
    \toprule
        Methods & Cora & CiteSeer & PubMed & Chameleon & Cornell & Actor & Squirrel & Texas & Wisconsin & Rank$\downarrow$  \\ \midrule
\model{} & \textbf{89.72 $\pm$ 1.46}& \textbf{79.30 $\pm$ 2.13}& 89.43 $\pm$ 0.53& \textbf{68.86 $\pm$ 1.45}& \textbf{80.00 $\pm$ 4.26}& \textbf{37.21 $\pm$ 1.35}& \textbf{54.78 $\pm$ 1.58}& \textbf{86.22 $\pm$ 4.67}& \textbf{85.88 $\pm$ 3.18}& \textbf{1.56} \\ 
~~~~w/o $A'$ & {87.83 $\pm$ 0.95}& 75.56 $\pm$ 1.32& \textbf{89.48 $\pm$ 0.58}& 67.94 $\pm$ 1.68& 77.57 $\pm$ 5.26& 35.09 $\pm$ 1.18& 53.35 $\pm$ 1.33& 82.70 $\pm$ 6.27& 83.53 $\pm$ 3.83& \underline{3.78} \\ 
~~~~w/o $A$ & 87.63 $\pm$ 1.29& 75.20 $\pm$ 1.79& 88.15 $\pm$ 0.65& 63.86 $\pm$ 1.42& 75.41 $\pm$ 5.76& 33.98 $\pm$ 1.20& 46.68 $\pm$ 1.32& 76.49 $\pm$ 5.56& 82.75 $\pm$ 4.87& 8.00 \\ 
~~~~$A + A'$ & 87.40 $\pm$ 1.34& 75.31 $\pm$ 1.93& 88.04 $\pm$ 0.64& 68.00 $\pm$ 1.87& 78.11 $\pm$ 5.17& 34.03 $\pm$ 1.25& 54.18 $\pm$ 1.33& 81.89 $\pm$ 4.60& {83.53 $\pm$ 5.16}& 5.06 \\ \bottomrule
    \end{tabular}}
    \label{tab:abla}
\end{table*}

\subsection{Hyper-parameter Analysis}
\begin{figure} 
    \centering
  \subfloat[]{%
       \includegraphics[width=0.5\textwidth]{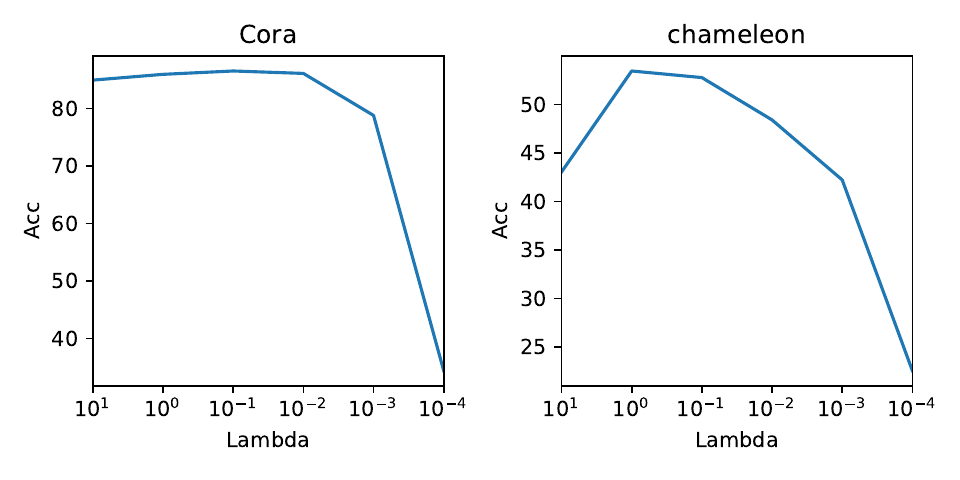}\label{fig:hyper_lambda}}
    \hfill
  \subfloat[]{%
        \includegraphics[width=0.5\textwidth]{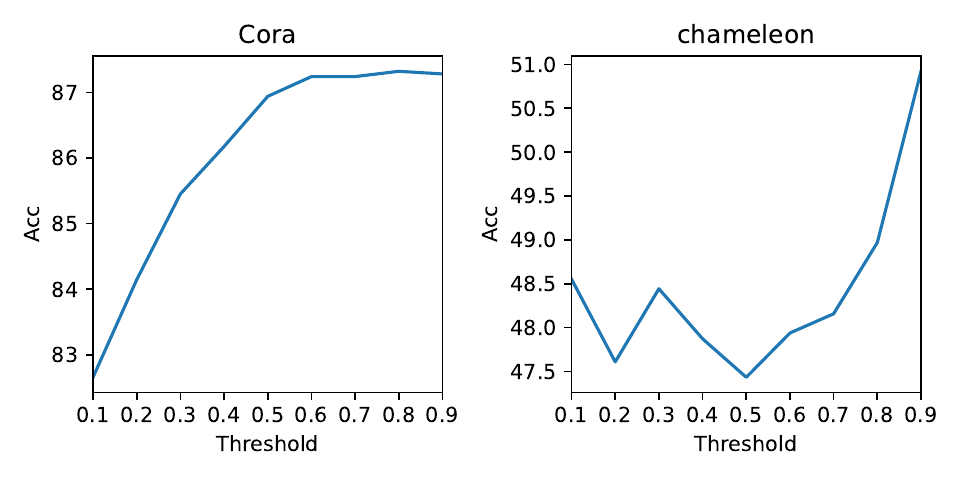}\label{fig:hyper_threshold}}
    \hfill
    \caption{The hyper-parameter analysis of lambda $\lambda$ and threshold $\delta$ in HiGNN on homophlious dataset(Cora) and heterophilous dataset(Chameleon)}
\end{figure}

This section introduces the influence of the threshold parameter $\delta$ and the lambda parameter $\lambda$ on \model{}. The $\delta$ modulates the similarity threshold of heterophilous information in constructing $A'$ and the $\lambda$ balances the significance of the new channel within the \model{} framework. As shown in Fig. \ref{fig:hyper_lambda} a smaller $\lambda$ leads to a decrease in model performance for both datasets, indicating the necessity of introducing heterophilous information. An overlarge $\lambda$ also leads to the degradation of the performance, especially for Chameleon. This implies the optimal $\lambda$ should be selected from $0.1$ to $1$. For the threshold $\delta$ as shown in Fig. \ref{fig:hyper_threshold}, the performance of \model{} becomes better with the increase of $\delta$. This shows a larger $\delta$ could capture more efficient connections between semantically similar nodes, which correlates with the Theorem \ref{theo:new_hom}.

In addition, we report the hyper-parameters for HiGNN. As shown in Table \ref{tab:hyperpara_hignn}, HiGNN is not sensitive to most of the basic hyper-parameters including learning rate, dropout size, lambda, number of layer. This indicates stable performance of HiGNN in general GNN settings. Regarding the threshold $\delta$, the best results are achieved with high values, suggesting that a higher similarity of heterophilous information is always preferable. As for the lambda $\lambda$, the best results vary across different datasets, indicating that an optimal lambda $\lambda$ should be selected for each graph.

\begin{table*}[!ht]
    \centering\caption{Hyper-parameters for HiGNN}
    \resizebox{0.9\hsize}{!}{
    \begin{tabular}{l|ccccccccc}
    \toprule
Hyper-parameter & Cora& Citeseer& Pubmed& Chameleon& Cornell& Film& Squirrel& Texas& Wisconsin\\ \midrule
Learning Rate & $10^{-3}$ & $10^{-3}$& $10^{-2}$& $10^{-2}$& $10^{-2}$& $10^{-2}$& $10^{-2}$& $10^{-2}$& $10^{-2}$\\
Dropout Rate& 0.4& 0.7& 0.4& 0.7& 0.6& 0.7& 0.6& 0.7& 0.7\\
Embedding Size& 128& 256& 128& 256& 128& 64& 256& 128& 256\\
Weight Decay& $5\times 10^{-5}$& $5\times 10^{-5}$& $5\times 10^{-5}$& $5\times 10^{-5}$& $5\times 10^{-5}$& $5\times 10^{-5}$& $5\times 10^{-5}$& $5\times 10^{-5}$& $5\times 10^{-5}$\\
Lambda & 1& 1& $10^{-2}$& $10^{-4}$& $5\times 10^{-2}$& 1& $5\times 10^{-2}$& $10^{-2}$& 0.01\\
Layer Number & 2& 2& 2& 2& 2& 2& 2& 2& 2\\
Threshold & 0.8& 0.9& 1.0& 0.8& 1.0& 0.8& 0.8& 0.9& 1.0\\ \toprule
    \end{tabular}}
    \label{tab:hyperpara_hignn}
\end{table*}
\section{Related Work}
\subsection{Graph Neural Networks}
Graph Neural Networks are effective for handling graph-structured data\cite{rw_gnn1, rw_gnn2}. These networks capture the dependencies between nodes through a message-passing mechanism. As a representative of GNNs, Graph Convolutional Networks (GCNs)\cite{GCN} perform convolution on the spectral domain to aggregate information from neighbors, capturing the relationships between nodes in graphs. Graph Attention Networks (GATs)\cite{GAT} specify different weights to different neighbors with attention mechanisms, enabling the model to focus on the most important information. For inductive representation learning, GraphSage\cite{graphSage} generates low-dimensional vectors by learning a function that aggregates neighbors. These GNNs operate effectively under the assumption of homophily, which posits that connected nodes tend to exhibit similarity. However, these methods experience significant performance degradation in heterophilous graphs, where connected nodes are more likely to have different labels\cite{rw_hom_asump}.

\subsection{Graph Homophily}
To solve the issue of graph heterophily, many methods have been proposed to improve graph representation learning. These methods can be categorized into two classes: aggregation calibration methods and neighbor extension methods. 

The aggregation calibration methods primarily focus on improving message aggregation. Mixhop\cite{Mixhop} integrates embeddings of k-hop neighbors during the graph convolution and employs trainable aggregation parameters for each hop. Following these high-order aggregations of neighbors, H2GCN\cite{H2GCN} further distinguishes between ego and neighbor embeddings and combines intermediate representations with theoretical justification. GPR-GNN\cite{GPR-GNN} uses a Generalized PageRank to learn weights for k-hop neighbor aggregations. HOG-GCN\cite{HOG-GCN} constructs a homophily degree matrix with attribute and topological information to adaptively modify the feature propagation process. To enhance the computational efficiency on large-scale graphs, LINKX\cite{LINKX} considers the adjacency matrix as an additional channel of node features. CPGNN\cite{CPGNN} learns a compatibility matrix by modeling the prior beliefs of the nodes. LFL\cite{leve_free} designs an adaptive filter initialized with link predtion under the weakly-supervised settings.

The neighbor extension methods focus on constructing new filters or new graph structures to extend neighbors to global nodes. To adaptively integrate different signals during the message aggregation, FAGCN\cite{FAGCN} incorporates both the low and high-frequency filters in GCNs. Based on this approach, ACM-GCN\cite{ACM-GCN} further employs low-pass, high-pass, and identity channels to adaptively mix different levels of frequency signals. To preserve topological information and capture long-range dependencies, Geom-GNN\cite{Geom-GNN} constructs structural neighbors using geometric measurements. To address both the over-smoothing and heterophily problems, GGCN\cite{GGCN} corrects original edges by considering node structures or features. To extend neighbors to nodes with similar semantics, WRGAT\cite{WRGAT} learns a new computation graph based on proximity and local structural similarity of nodes. To aggregate the information from global nodes in graphs, GloGNN\cite{GloGNN} learns a coefficient matrix to capture the correlations between nodes by considering both feature similarity and topology similarity. L2A\cite{L2A} performs graph structure augmentation into a continuous optimization problem with a variational inference approach. 

The aggregation calibration methods perform message passing from a local perspective, thereby failing to capture long-range dependencies from global nodes. To address the issue, current neighbor extension methods establish new connections with global nodes by utilizing either feature information, node structural information, or new filters. However, these methods overlook heterophilous information, which describes the contexts of homophilous or heterophilous nodes. This heterophilous information provides rich semantics of a node that cannot be obtained through node features or topological information.
\section{Conclusion}

This paper introduces the concept of heterophilous information as the distribution of node neighbors and proposes a practical solution to effectively leverage the rich semantics present in graphs. In other words, the heterophily graph structure is regarded as semantic information rather than noise information. HiGNN is proposed to make good use of the semantics to improve the connectivity between nodes with identical labels, thereby accommodating datasets with varying homophily, as shown by our theoretical and empirical analysis. Our experiments show the superiority of \model{} in handling both homophilous and heterophilous graphs. It is worth noting that while \model{} enhances the graph learning, it depends on the prior estimation of all node labels. In our future work, we aim to explore more efficient representations of heterophilous information that strike a balance between computational demands and performance.

\ifCLASSOPTIONcaptionsoff
  \newpage
\fi



%
\bibliographystyle{IEEEtran}
\bibliography{ref}

%

\begin{IEEEbiography}[{\includegraphics[width=1in,height=1.25in,clip,keepaspectratio]{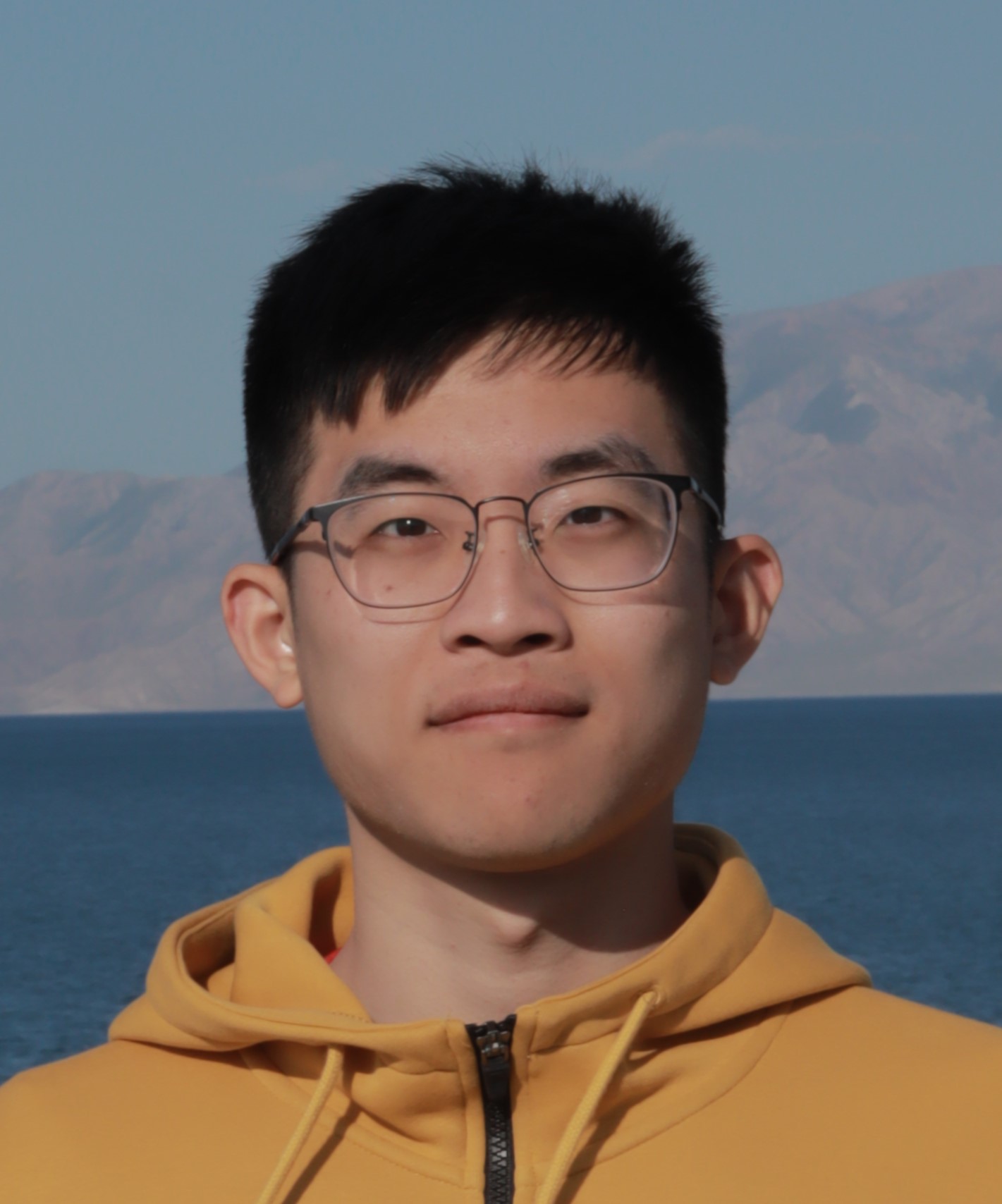}}]{Zheng Yilun} is currently a Ph.D. candidate at Nanyang Technological University supervised by prof. Chen Lihui. He received the Bachelor's degree from Zhejiang University of Technology, Hangzhou, China in 2022. His research interests include graph representation learning, graph homophily and natural language processing.
\end{IEEEbiography}

\begin{IEEEbiography}[{\includegraphics[width=1in,height=1.25in,clip,keepaspectratio]{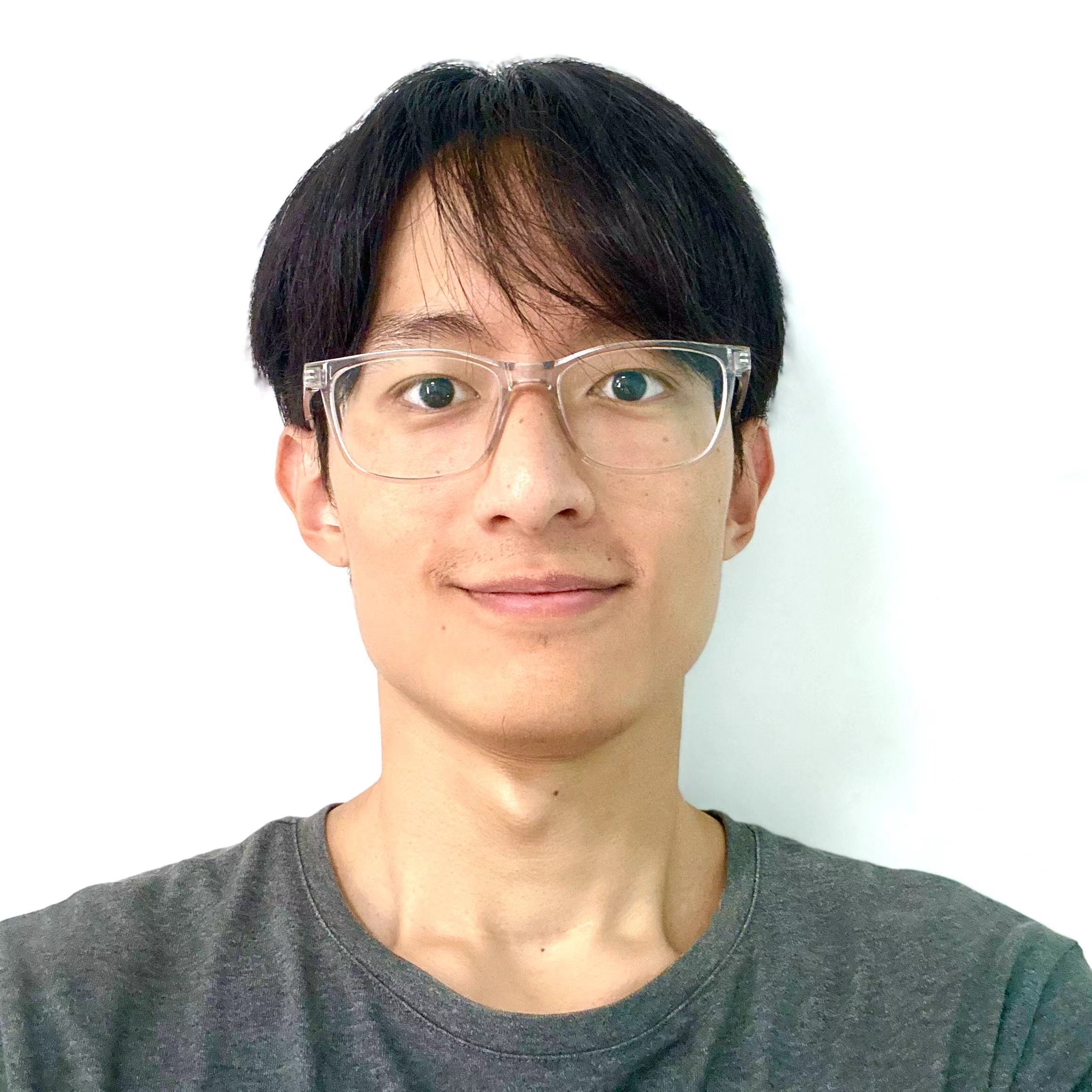}}]{Jiahao Xu} is currently a Ph.D. candidate at Nanyang Technological University. Before that, he received the master’s degree in signal processing at the same university at 2018 and  B.Eng. degree in School of Electrical and Electronic Engineering from University of Electronic Science and Technology of China (UESTC), Sichuan, China in 2018. His research interests include semantic sentence representation, neural machine translation, knowledge graph natural language processing, and graph representation learning.
\end{IEEEbiography}

\begin{IEEEbiography}[{\includegraphics[width=1in,height=1.25in,clip,keepaspectratio]{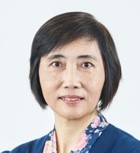}}]{Lihui Chen}
(SM’07) received the B.Eng. degree in computer science and engineering from Zhejiang University, Hangzhou, China, and the Ph.D. degree in computational science from the University of St Andrews, St Andrews, U.K. She is currently an Associate Professor with the Centre for Information Sciences and Systems, School of Electrical and Electronic Engineering, Nanyang Technological University, Republic of Singapore. She has authored over 110 referred papers in international journals and conferences in her research areas. Her current research interests include machine learning algorithms and applications, graph data analytics, text/data mining, data clustering, and deep representation learning.
\end{IEEEbiography}








\end{document}